\documentclass[10pt,twocolumn,letterpaper]{article}

\usepackage{wacv}
\usepackage{times}
\usepackage{epsfig}
\usepackage{graphicx}
\usepackage{amsmath}
\usepackage{amssymb}
\usepackage{multirow}

\wacvfinalcopy 

\def\wacvPaperID{1064}

\ifwacvfinal\fi
\begin{document}

%%%%%%%%% TITLE
\title{ParkingSticker: A Real-World Object Detection Dataset}

% % Authors at the same institution
% \author{Ethem F. Can \hspace{0.8cm} Caroline Potts \hspace{0.8cm} 
% Aysu Ezen Can \hspace{0.8cm}Xiangqian Hu \hspace{0.8cm} \\
% SAS\\
% {\tt\small \{caroline.potts, ethem.can, xiangqian.hu, aysu.ezencan, saratendu.sethi\}@sas.com}
% }
% Authors at different institutions
\author{Caroline Potts \\
{\tt\small caroline.potts@duke.edu}
\and
Ethem F. Can \\
{\tt\small ethfcan@gmail.com}
\and
Aysu Ezen-Can \\
{\tt\small aysu.e.can@gmail.com}
\and
Xiangqian Hu \\
{\tt\small xiangqian.hu@sas.com}
}

\maketitle
\ifwacvfinal\thispagestyle{empty}\fi

%%%%%%%%% ABSTRACT
\begin{abstract}
   We present a new and challenging object detection dataset, ParkingSticker, which mimics the type of data available in industry problems more closely than popular existing datasets like PASCAL VOC.  ParkingSticker contains 1,871 images that come from a security camera's video footage.  

The objective is to identify parking stickers on cars approaching a gate that the security camera faces.  Bounding boxes are drawn around parking stickers in the images. The parking stickers are much smaller on average than the objects in other popular object detection datasets; this makes ParkingSticker a challenging test for object detection methods.  This dataset also very realistically represents the data available in many industry problems where a customer presents a few video frames and asks for a solution to a very difficult problem.  

Performance of various object detection pipelines using a YOLOv2 architecture are presented and indicate that identifying the parking stickers in ParkingSticker is challenging yet feasible.  We believe that this dataset will challenge researchers to solve a real-world problem with real-world constraints such as non-ideal camera positioning and small object-size-to-image-size ratios. 
\end{abstract}

%%%%%%%%% BODY TEXT
\section{Introduction}

Realistic datasets are extremely important in driving and measuring advances in object detection, especially in industry settings. To achieve goals of using object detection to solve challenging real-world problems, real-world data is needed for testing and improving systems. Existing image datasets serve as useful benchmarks for testing many computer vision systems and provide large amounts of labeled data.  In industry problems, however, obtaining and using such large amounts of data is often too expensive, both in collection and computation costs, so these large datasets do not provide a full understanding of how models will perform.  Additionally, some popular existing datasets like ImageNet contain the object of interest in the center of most of the images, which is not realistic in many industry cases.  

This work introduces a new dataset, ParkingSticker.  ParkingSticker represents the data that one might receive in an industry project more closely than existing object detection datasets because it is relatively small, the images in it come from videos, and the lighting conditions and camera positioning are not ideal.  Changing these conditions was not an option in the collection of this data, and it is often not an option in industry problems.  ParkingSticker also differs from existing datasets in that the objects to be detected in it (parking stickers) are very small.  

In addition to presenting the dataset, we present preliminary experimental results using ParkingSticker.  These results indicate that the detection of the parking stickers in the images is a feasible but challenging problem, one that leaves room for future improvements.

%------------------------------------------------------------------------
\section{Related work}

The Modified National Institute of Standards and Technology (MNIST) dataset \cite{MNIST} contains 60,000 training and 10,000 test images of handwritten digits.  The digits are normalized and centered in 28 x 28 pixel images.  MNIST has become a standard dataset for testing image classification algorithms.  It is, however, unlike many real-world problems because the data is so standardized.  The images are also much smaller than those from modern cameras.  In contrast, the images from ParkingSticker contain the object of interest (the parking stickers) in various locations in the images (see Figure \ref{fig: sticker locations}) and are originally 1920 x 1080 pixels.  

The CIFAR-10 dataset \cite{CIFAR}, like MNIST, contains 10 classes.  The images in CIFAR-10 are a subset of the 80 million tiny images dataset \cite{80mil}.  The CIFAR-100 dataset \cite{CIFAR} is also a labeled subset of the 80 million tiny images dataset, but it contains 100 object classes instead of only 10. Both datasets contain 60,000 images, split evenly among the classes.  Both are useful datasets for benchmarking image classification techniques for research purposes, but they are too perfectly crafted to capture how techniques will perform for an industry problem.  

When it was introduced in 2009, ImageNet \cite{ImageNet} represented a huge jump in dataset size over MNIST, CIFAR-10, and CIFAR-100; it now contains over 14 million labeled images in over 21,000 classes.  Because of its size and the popular ImageNet Large Scale Visual Recognition Challenge, ImageNet is a very commonly used dataset for image classification.  The ImageNet Object Localization Challenge is a competition dataset that contains 150,000 images with objects in 1,000 categories annotated with bounding boxes, enabling ImageNet to also be used for object detection research. ImageNet's large size makes it extremely useful in setting initial weights of models for transfer learning tasks.  It does not, however, mimic the real-world conditions that created ParkingSticker; the images were collected through internet image searches rather than through one specific real-world camera.  ImageNet images tend to have one large, centered object of interest.  In contrast, the objects annotated in ParkingSticker are much smaller than most of those in ImageNet and appear in various locations in the images, as is described further in Section \ref{dataset}. The images in ParkingSticker are also higher resolution than those in ImageNet; the average resolution of ImageNet images is 400 x 350. 

While ImageNet can be used in both classification and localization tasks, the PASCAL VOC dataset \cite{PASCAL} is intended specifically for use in localization problems.  It contains 11,000 images with objects identified in 20 categories.  It includes over 27,000 bounding box annotations, making it a popular choice for object detection research.  Around 7,000 of these object instances also have detailed segmentations.  Like ImageNet, PASCAL VOC contains images that are less representative of a real-world industry case than those in ParkingSticker.  

Another recent object detection dataset is Microsoft COCO (Common Objects in Context) \cite{COCO}.  COCO contains 328,000 images with 2.5 million labeled objects from 91 categories.  Both segmentations of the objects and bounding boxes around them are available.  The images show objects in their normal contexts, rather than iconic object or scene images.  ParkingSticker similarly shows parking stickers in the context in which they would commonly be seen by security cameras: on the windshields of approaching cars.  Like the other previously mentioned datasets, COCO does not contain many sequential images taken from videos; COCO contains at most five images taken by any single photographer in a short period of time.  The bounding boxes in COCO are also much larger on average than those in ParkingSticker (see Section \ref{dataset}).  

While these existing datasets are extremely useful tools for measuring object detection advances, ParkingSticker better simulates a common customer problem in which the settings like camera positioning are non-ideal; it therefore provides an opportunity for researchers to apply their methods to a very realistic dataset.  Accurate detection on this dataset is also an open problem that could quickly provide real-world value; it would allow the presence of parking stickers on cars to be checked before the cars enter a company's campus.  

%------------------------------------------------------------------------
\section{Dataset details}
\label{dataset}
\begin{figure*}[t]
\begin{center}
   \includegraphics[width=0.9\linewidth]{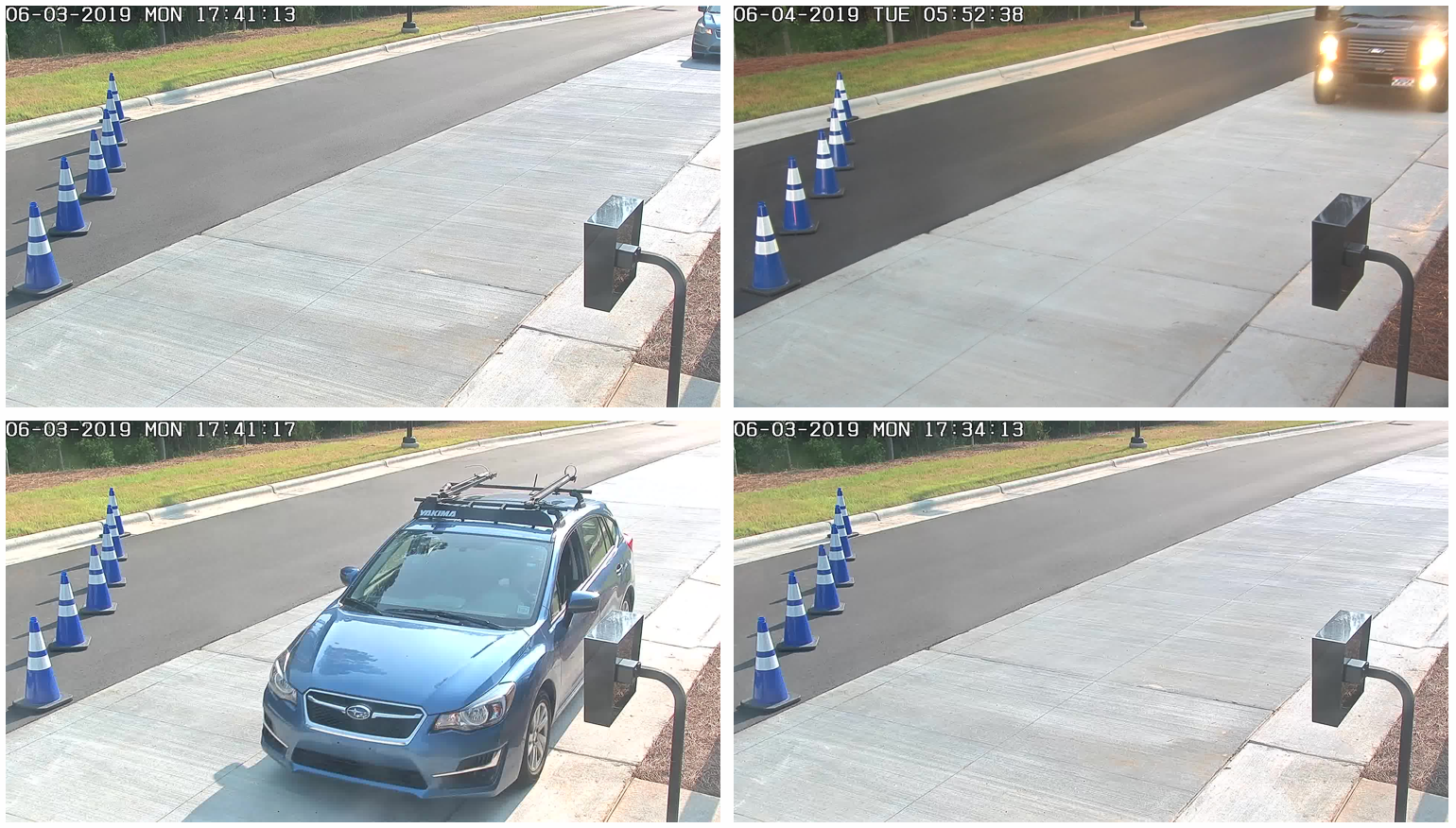}
\end{center}
   \caption{Sample frames from security camera videos.}
\label{fig: sample images}
\end{figure*}

Videos from a security camera at the gated entrances to a company's headquarters were collected over the span of several daylight hours.  The security camera recorded footage when there was movement detected in the surrounding area.  This movement was often from approaching vehicles but could also be from other sources including wind, animals, and people. Although the camera was usually pointed in the same direction, facing cars approaching the gate, it could and did sometimes move to point in other directions.  The videos were split into 29,348 images.  The images are HD quality, 1920 x 1080 pixels.  Figure \ref{fig: sample images} shows four sample images.  

We did our best to remove personally identifying information like faces and license plate numbers from the images.  
%-------------------------------------------------------------------------
\subsection{Removing images without vehicles}

Examination of the data revealed that many images did not contain any vehicles.  Images without vehicles necessarily also do not include any parking stickers.  These images, therefore, did not need to be annotated.  To reduce the time needed to annotate the data, all images were passed through a YOLOv2 object detector \cite{YOLOv2} trained on a subset of the Open Images V4 dataset \cite{OpenImages}.  The images were resized to 416 x 416 before being passed through this pretrained object detector.  This object detector identified whether a vehicle was present in each image.  The images without vehicles were removed from the dataset.  After removing the images without detected vehicles, the dataset contained 4,253 HD images.  

Because the pretrained object detector does not have perfect accuracy, this method does result in some images with vehicles being removed from the dataset.  This loss in data was determined to be small enough to be neglected.  

\subsection{Data annotation}

\begin{figure}[t]
\begin{center}
   \includegraphics[width=0.8\linewidth]{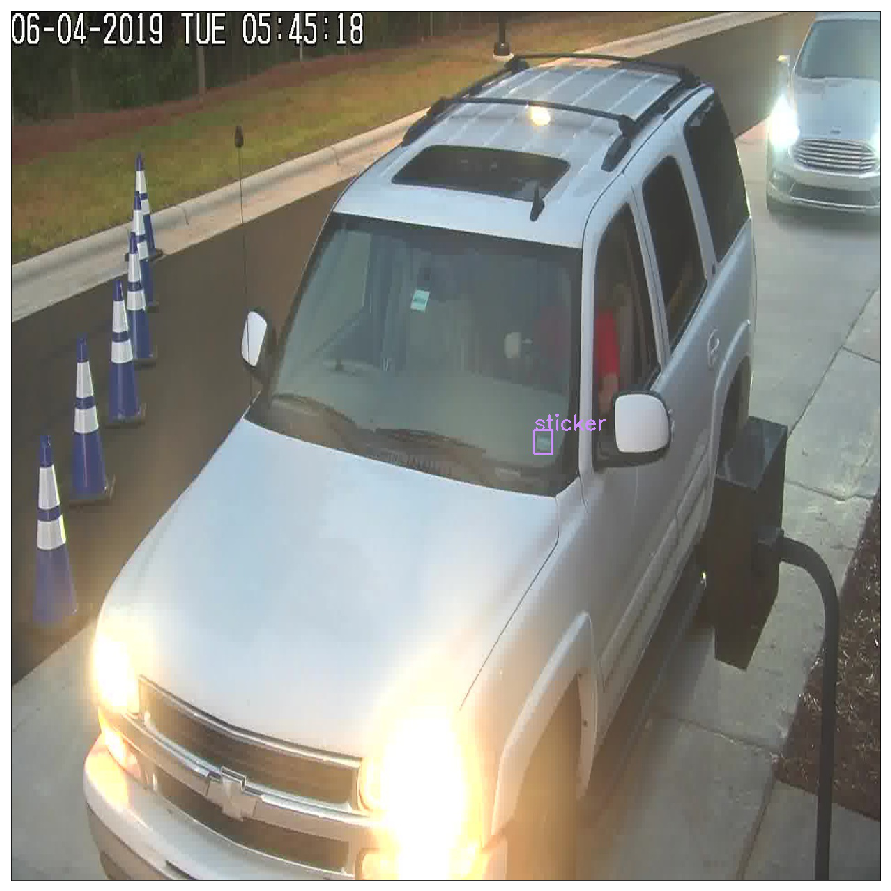}
\end{center}
   \caption{Sample annotated image, resized to be square.}
\label{fig: annotated image}
\end{figure}

To annotate the images in which vehicles were detected, a custom tool was used to draw bounding boxes around the parking stickers.  The coordinates of the boxes were saved in YOLO format.  YOLO format holds the bounding box (\textit{x, y, w, h}) coordinates.  The \textit{x} and \textit{y} coordinates are between 0 and 1; they are how far the top left corner of the bounding box is from the top left corner of the image, as a fraction of image width and height, respectively.  The \textit{w} and \textit{h} coordinates are the width and height of the bounding box as a fraction of the image width and height, respectively. 846 images contained one sticker, and no images contained more than one sticker.  A sample annotated image is shown in Figure \ref{fig: annotated image}. Because the dataset is so challenging, even annotation of the stickers was difficult.  In some images, it was hard to tell if the car had a parking sticker or not.  

\subsection{Dataset statistics}
\begin{figure}[t]
\begin{center}
   \includegraphics[width=0.8\linewidth]{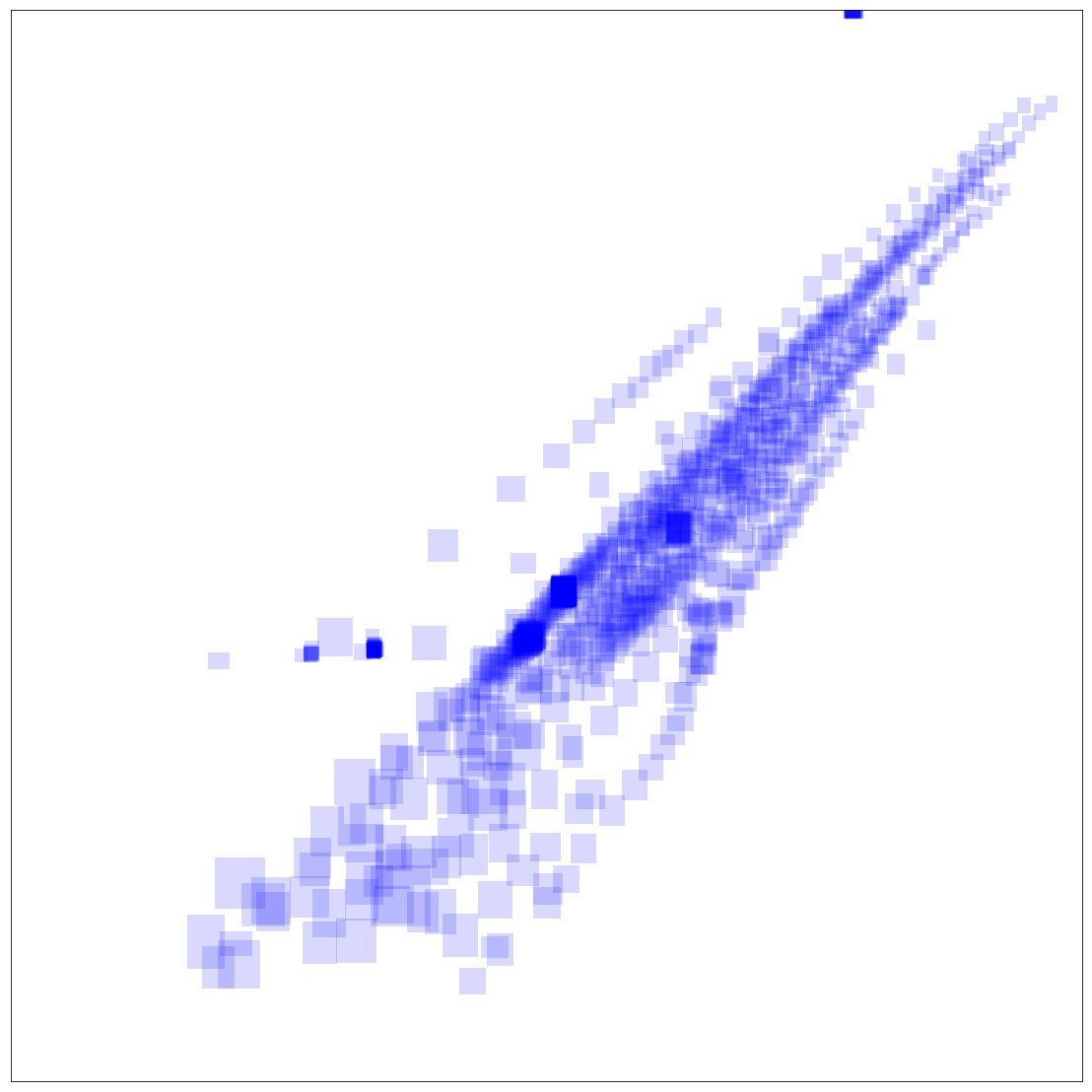}
\end{center}
   \caption{Parking sticker locations and sizes in images.  The size and location of the rectangles relative to the figure window is equal to the size and location of the bounding boxes relative to the images.  }
\label{fig: sticker locations}
\end{figure}

Figure \ref{fig: sticker locations} shows the locations and relative sizes of the parking sticker bounding boxes.  The parking stickers tend to be located along a diagonal line from the top right corner to the bottom left corner.  They tend to increase in size as they move along the line towards the bottom left corner.  This pattern is due to the camera setup; the camera was usually pointed in the same direction and captured many frames of each sticker as the car moved towards it.  Notably, not all sticker bounding boxes fall along this line because the camera was not entirely stationary. 

\begin{figure}[t]
\begin{center}
   \includegraphics[width=0.9\linewidth]{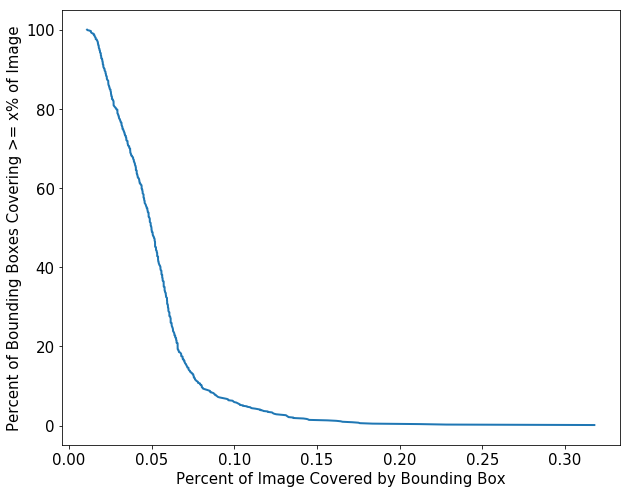}
\end{center}
   \caption{Percentage of bounding boxes in ParkingSticker covering at least various percentages of images.}
\label{fig: sticker sizes}
\end{figure}

As can be seen in Figure \ref{fig: sticker locations}, the bounding boxes are quite small compared to the size of the images.  Figure \ref{fig: sticker sizes} shows the percentage of bounding boxes covering at least various percentages of the images.  The average bounding box covers 0.052\% of the image area.  The maximum percentage of area covered by one box is 0.32\%, and the minimum percentage is 0.011\%.  

The bounding boxes in ParkingSticker are much smaller than those in standard existing datasets.  In ImageNet, COCO, and PASCAL VOC, only approximately 29\%, 20\%, and 18\% of images respectively cover less than 4\% of the image area \cite{COCO}.  

%-------------------------------------------------------------------------
\section{Experimental setup}
This section describes our experimental setup for the sticker detection task using a variation of the YOLOv2 architecture.  

\subsection{Splitting data into sets}

The 846 images with stickers were divided in an approximately 70-10-20 split into training, validation, and test sets.  Images containing cars without stickers were added to the validation and test sets; the ratio of images with stickers to those without sticker is approximately equal to the ratio in the entire set of images containing vehicles.  Table \ref{tbl: set info} shows the number of images with and without stickers in each set.

\begin{table}
\begin{center}
\begin{tabular}{|c|c|c|c|}
\hline
\begin{tabular}[c]{@{}c@{}}Set\end{tabular} &  \begin{tabular}[c]{@{}c@{}}Images\\with\\Stickers\end{tabular}
&\begin{tabular}[c]{@{}c@{}}Images\\without\\Stickers\end{tabular}&\begin{tabular}[c]{@{}c@{}}Total\\Number of\\Images\end{tabular}\\
\hline
\hline
Training & 587 & 0 & 587\\
\hline
Validation & 86 & 344 & 430 \\
\hline
Test & 173 & 681 & 854\\
\hline 
\end{tabular}
\end{center}
\caption{Number of images in each set.}
\label{tbl: set info}
\end{table}

All of the sets contain only images with cars.  Because vehicle detection is an easier problem to solve than parking sticker detection based on the relative sizes of the two objects, any images from the security camera footage would first be passed through a vehicle detector.  Only if there is a vehicle detected in the image will the parking sticker detector be used; the parking sticker detector, therefore, should only be trained and tested on images containing vehicles.  Images without parking stickers but with cars were added to the validation and test sets because in a real-world case, not all cars approaching the security cameras will have a parking sticker anywhere on them.  Differentiating between cars with and without stickers is crucial for this use case, so the validation and test sets need to reflect both cases.  

\subsection{Evaluation metrics}

While object detection was used instead of binary image classification since the small size of the stickers would make this a very challenging binary image classification problem, the exact location of the sticker is of less interest than its presence or absence.  The intersection over union (IOU), a commonly used object detection metric, measures the accuracy of the detection bounding boxes.  We also used precision, recall, and F1 score to evaluate how accurately we were detecting the stickers.

\subsection{Experiment details}

A variation of the YOLOv2 architecture \cite{YOLOv2} was used to detect the parking stickers. In addition to the original YOLOv2 architecture, the model adds a passthrough layer that brings features from an earlier layer to lower resolution layer. This architecture is fast and lightweight, making it suitable for real-time parking sticker detection.  

Input image sizes of 416 x 416 and other multiples of 416 up to 1664 x 1664 were tried.  A batch size of 16 and Adam optimization were used.  The models were trained for 100 epochs, for which the learning rate policy was as follows: 40 epochs with learning rate = 0.001, 40 epochs with learning rate = 0.0001, 20 epochs with learning rate = 0.00001.  

The YOLOv2 loss function contains four hyperparameters, each controlling how severely different types of mistakes are penalized.  One hyperparameter controls the impact of misclassifying a detected object, if one is detected.  Another controls the impact of predicted bounding box coordinates that do not align well with the ground truth coordinates.  Both of these were left at their default values of 0.5 and 5, respectively.  The other two, which we will call the ``confidence loss object hyperparameter" and the ``confidence loss no object hyperparameter," control the penalty for determining that an image section does not contain an object when it does, and vice versa.  Because the parking stickers are so small, the image contains overwhelmingly more ``no object" sections than ``object" sections.  To address this imbalance, the confidence loss object hyperparameter was set to 35, and the confidence loss no object hyperparameter was left at its default value of 0.5.  The value 35 and the other model parameters discussed in this section were chosen based on experiments with trying to overfit to the training data.  

To increase the variation in the dataset and decrease overfitting, random mutations of the images were done before every model was trained.  These random mutations consisted of color jittering, color shifting, sharpening, lightening, and darkening.  

All experiments were run using the SAS Viya 3.4 platform with VDMML 8.4 on the backend.  DLPy, a deep learning Python API, was used on the frontend. 

%-------------------------------------------------------------------------
\section{Results and discussion}

The most common input image size for YOLOv2 is 416 x 416 \cite{YOLOv2}.  Because the parking stickers are on average only 0.052\% of the image area, smaller than in most use cases, it was hypothesized that a higher resolution would improve results.  The impact that increasing the image size has on the evaluation metrics is shown in Table \ref{tbl: size metrics}.  As can be seen in the table, increasing input image resolution consistently increased precision and F1 score.  Recall, however, was highest when the input images were 1248 x 1248.  In the 1248 x 1248 images, almost all true stickers were found, along with a number of false positives.  

\begin{table}
\begin{center}
\begin{tabular}{|c|c|c|c|}
\hline
   \begin{tabular}[c]{@{}c@{}}Image Size\end{tabular} &  \begin{tabular}[c]{@{}c@{}}Precision\end{tabular}
   &\begin{tabular}[c]{@{}c@{}}Recall\end{tabular}&\begin{tabular}[c]{@{}c@{}}F1 Score\end{tabular}\\
\hline
\hline
416 x 416 & 0.1972 & 0.6512 &  0.3027 \\
\hline
832 x 832 & 0.2555 & 0.9535 & 0.4029\\
\hline
1248 x 1248 & 0.3172 & 0.9884& 0.4802 \\
\hline
1664 x 1664 & 0.6140 & 0.8140 & 0.7000\\
\hline 
\end{tabular}
\end{center}
\caption{Effects of input image size on evaluation metrics. All metrics come from YOLOv2 models used on the validation set.  }
\label{tbl: size metrics}
\end{table}

Increasing input image resolution also increases the time required for both training and inference, as is shown in Table \ref{tbl: size efficiency}.  All experiments shown in the table were run on a P100 GPU card with 16GB of memory.  A minibatch size of 16 was used.  Since the detection of parking stickers is most useful if done in real-time, keeping the inference time low is very important in this application.  

\begin{table}
\begin{center}
\begin{tabular}{|c|c|c|}
\hline
   \begin{tabular}[c]{@{}c@{}}Image \\ Size\end{tabular} &\begin{tabular}[c]{@{}c@{}}Training \\ Time\\ (100 epochs)\end{tabular}&\begin{tabular}[c]{@{}c@{}}Inference \\ Time\\ (430 images)\end{tabular}\\
\hline
\hline
416 x 416 &  1635.95s & 7.98s\\
\hline
832 x 832 & 4282.53s & 12.48s\\
\hline
1248 x 1248 & 11586.16s & 23.38s\\
\hline
1664 x 1664 & 22143.49s & 40.09s\\
\hline 
\end{tabular}
\end{center}
\caption{Effects of input image size on efficiency of training and testing YOLOv2 models.  }
\label{tbl: size efficiency}
\end{table}

While increasing the input image size improved the accuracy of the parking sticker detection, there was still significant room for improvement.  In particular, precision was fairly low for all input image sizes.  To boost performance further, two additions to the detection pipeline were tried: keeping only one detection per image and a binary classification step. These two additions will be explained in the following subsections.

\subsection{Keeping one detection per image}

\begin{figure}[t]
\begin{center}
   \includegraphics[width=0.9\linewidth]{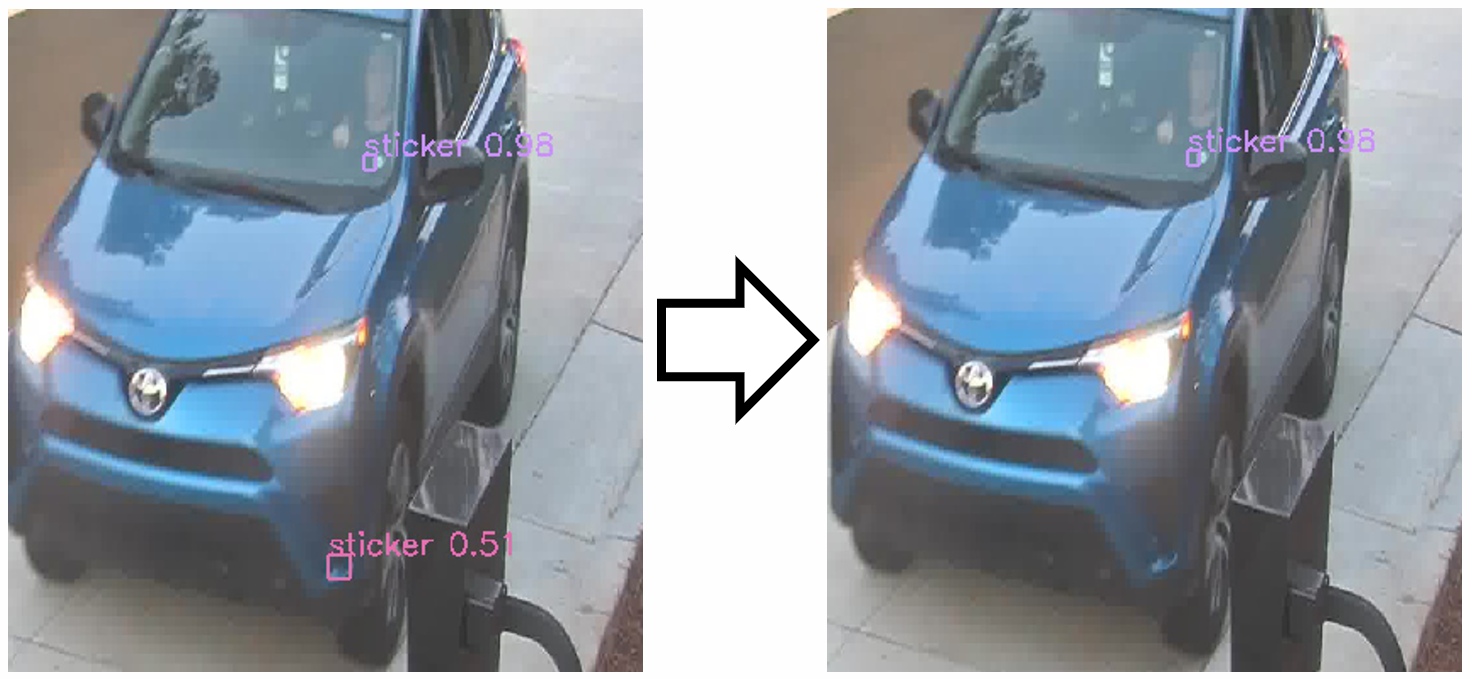}
\end{center}
\caption{Removal of a detected sticker with lower confidence than another detected sticker in the same image.}
\label{fig: keep one}
\end{figure}

Examining the predicted sticker locations from using all four input image sizes revealed that in some images, multiple stickers were detected.  However, there were no images that contained more than one true sticker annotation.  To address this, only the predicted parking sticker box with the highest confidence was kept in each image.  All other predicted boxes were discarded.  Figure \ref{fig: keep one} shows an example of an image where this technique had the desired effect; the true positive was kept while the false positive was discarded.  

The impact of this technique on the evaluation metrics is shown in Table \ref{tbl: one det metrics}.  Keeping only the detection with the highest confidence in each image consistently increased precision slightly while leaving recall unchanged, leading to a slight increase in F1 score. The slight increase in precision without any drop in recall indicates that this technique eliminated some false positives while leaving the true positives untouched.  These results indicate that in images in which more than one sticker is predicted, the true sticker tends to have the highest confidence value. 

\begin{table}
\begin{center}
\begin{tabular}{|c|c|c|}
\hline
\begin{tabular}[c]{@{}c@{}}Image \\ Size\end{tabular} &  \begin{tabular}[c]{@{}c@{}}Keeping All\\ Detections\end{tabular} & \begin{tabular}[c]{@{}c@{}}Keeping One\\ Detection\\ Per Image\end{tabular} \\
\hline
\hline
416 x 416 & \begin{tabular}[c]{@{}c@{}}Precision: 0.1972\\ Recall: 0.6512 \\ F1 score: 0.3027\end{tabular} &                     \begin{tabular}[c]{@{}c@{}}Precision: 0.2007\\ Recall: 0.6512\\ F1 score: 0.3068\end{tabular} \\
\hline
832 x 832 &\begin{tabular}[c]{@{}c@{}}Precision: 0.2555\\ Recall: 0.9535\\ F1 score: 0.4029\end{tabular}&                            \begin{tabular}[c]{@{}c@{}}Precision: 0.2733\\ Recall: 0.9535\\ F1 score:  0.4249\end{tabular} \\
\hline
1248 x 1248 &\begin{tabular}[c]{@{}c@{}}Precision: 0.3172\\ Recall: 0.9884\\ F1 score: 0.4802\end{tabular}&                       \begin{tabular}[c]{@{}c@{}}Precision: 0.3295\\ Recall: 0.9884\\ F1 score: 0.4942\end{tabular} \\
\hline
1664 x 1664 &\begin{tabular}[c]{@{}c@{}}Precision: 0.6140\\ Recall: 0.8140\\ F1 score: 0.7000\end{tabular}&                       \begin{tabular}[c]{@{}c@{}}Precision: 0.6250\\ Recall: 0.8140\\ F1 score: 0.7071\end{tabular} \\
\hline 
\end{tabular}
\end{center}
\caption{Effects of keeping only the predicted sticker with the highest confidence in each image on evaluation metrics. All metrics come from YOLOv2 models used on the validation set.  }
\label{tbl: one det metrics}
\end{table}

\subsection{Binary classification of predicted stickers}

The YOLOv2 object detectors for all image sizes resulted in higher recall than precision.  While many of the true stickers were detected, there were also many false positives. 

\begin{figure}[t]
\begin{center}
   \includegraphics[width=0.8\linewidth]{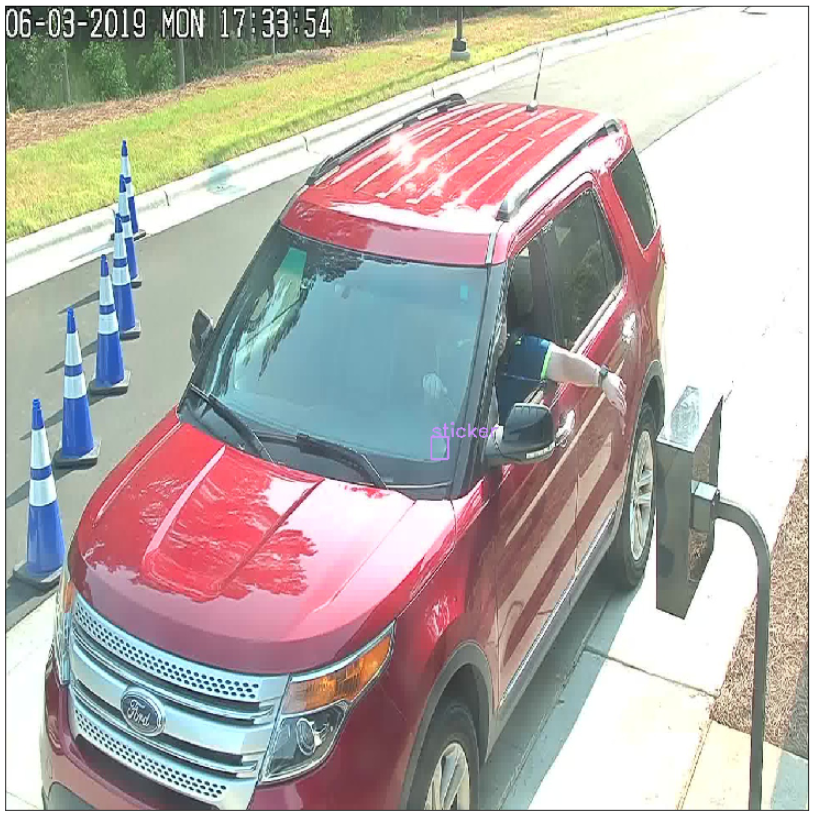}
\end{center}
\caption{False positive in the lower windshield corner on the driver's side, the place on cars where most true parking stickers were located.}
\label{fig: FP bottom corner}
\end{figure}

Inspection of the outputs of the object detection model revealed that many of the false positives appeared in the same location: the lower corner of cars' windshields on the driver's side (Figure \ref{fig: FP bottom corner}).  This location is where almost all of the true parking stickers were located; the object detection model, then, seem to have learned to identify not only parking stickers but also that specific windshield location.  

Ideally, more data would be collected with more diverse locations of parking stickers to address this phenomenon.  Collecting more data, however, is often not feasible in an industry setting; doing so is costly, and the customer may be unwilling or unable to provide more data.  

\begin{figure*}[t]
\begin{center}
   \includegraphics[width=0.99\linewidth]{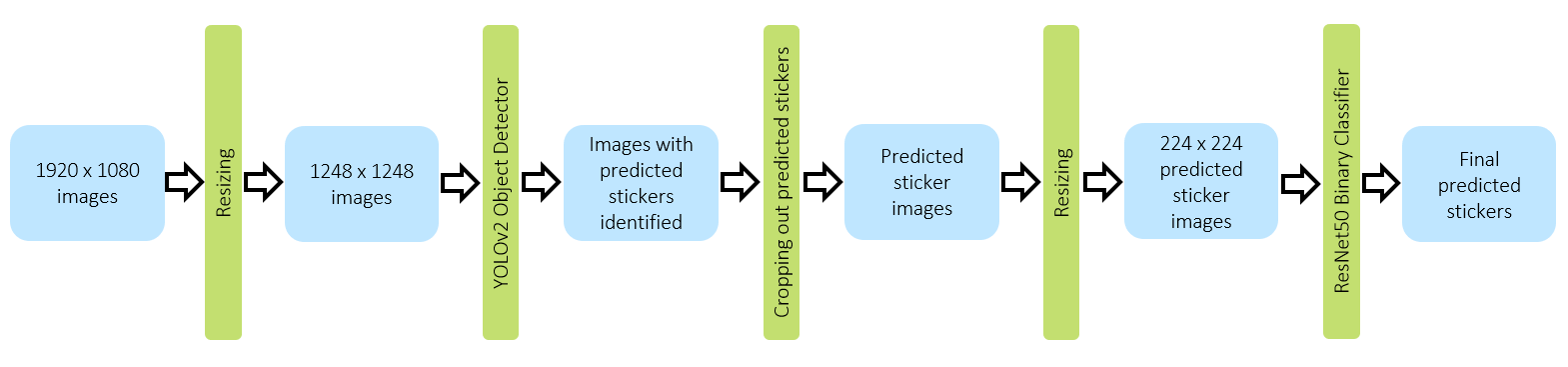}
\end{center}
   \caption{Sticker detection pipeline for 1248 x 1248 images with added binary classification step.  The predicted stickers output by the YOLOv2 object detector are only the predictions with the highest confidence in each image.  }
\label{fig: framework}
\end{figure*}

A different approach was therefore taken to address this issue; a binary image classification was done following the object detection on the 1248 x 1248 size images.  A diagram of the pipeline including the binary classification is shown in Figure \ref{fig: framework}.  The 1248 x 1248 size images were chosen for this task because that object detector resulted in the highest recall.  When the binary classification step is added to the prediction pipeline, recall is of primary importance in the object detection step.  False positives can be removed in the binary classification step; false negatives cannot.  

To perform the binary classification, the predicted detected stickers were first cropped out of the original images.  This removed the context of the rest of the image; it was hypothesized that this would help limit the number of false positives located on the lower driver's side windshield corners.  These cropped images were then passed through a ResNet50 image classifier \cite{ResNet50} that classified them as belonging to either the `sticker' or the `not sticker' class.  

\begin{figure}[t]
\begin{center}
   \includegraphics[width=0.95\linewidth]{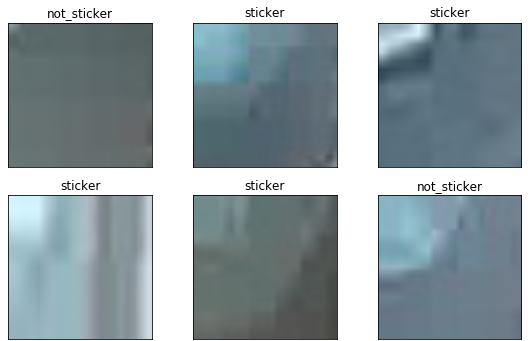}
\end{center}
   \caption{Examples of training images for ResNet50 binary classifier.  Images are cropped versions of the images in the extended training set.  They are cropped to contain only the pixels in the predicted sticker bounding boxes output by the YOLOv2 object detector and then resized to 224 x 224 pixels.}
\label{fig: cropped stickers}
\end{figure}

To train the ResNet50 classifier, a new extended training set of images was created.  This set consisted of the original 587 training images that contained annotations, plus 587 images that contained cars without stickers.  These additional 587 images were not included in either the validation or test sets.  The classifier was then trained using images cropped to the locations of the bounding boxes predicted by the object detector on the extended training set.  Sample images, resized to 224 x 224 as they were for the input to the binary classifier, are shown in Figure \ref{fig: cropped stickers}.  The images without stickers were added to increase the number of false positives output by the object detector and therefore the number of `not sticker' examples.  The ResNet50 classifier was trained for 60 epochs with a learning rate of 0.001, with initial weights set from training on the ImageNet images from the 1,000 classes that are included in the ImageNet Large Scale Visual Recognition Challenge \cite{ImageNet}.  

A comparison of the evaluation metrics of the system with different pipelines from 1248 x 1248 size images to final predictions is shown in Table \ref{tbl: pipeline comp}.  As the table shows, adding the binary image classifier to the pipeline increases precision substantially and causes a much smaller drop in recall; the F1 score therefore increases substantially. 

\begin{table}
\begin{center}
\begin{tabular}{|c|c|c|c|}
\hline
   \begin{tabular}[c]{@{}c@{}}Prediction \\Pipeline\end{tabular} &  \begin{tabular}[c]{@{}c@{}}Precision\end{tabular}
   &\begin{tabular}[c]{@{}c@{}}Recall\end{tabular}&\begin{tabular}[c]{@{}c@{}}F1 Score\end{tabular}\\
\hline
\hline
\begin{tabular}[c]{@{}c@{}}YOLOv2 object \\detector only\end{tabular} & 0.3172 & 0.9884& 0.4802 \\
\hline
\begin{tabular}[c]{@{}c@{}}YOLOv2 object \\detector \\+ \\Keeping only \\ highest confidence \\predictions\end{tabular} & 0.3295 & 0.9884 & 0.4942 \\
\hline
\begin{tabular}[c]{@{}c@{}}YOLOv2 object \\detector \\+ \\Keeping only \\ highest confidence \\predictions\\ + \\ Sticker vs. not \\sticker classifier\end{tabular} & 0.5900 & 0.9196 & 0.7182\\
\hline 
\end{tabular}
\end{center}
\caption{Comparison of evaluation metrics on validation set using different pipelines from 1248 x 1248 size images to final predictions.}
\label{tbl: pipeline comp}
\end{table}

\subsection{Recommendation}

The best results on the validation set were obtained using two methods: 
\begin{enumerate}
    \item 1664 x 1664 images: YOLOv2 object detector  +  keeping only the predicted stickers with the highest confidence in each image
    \item 1248 x 1248 images: YOLOv2 object detector  +  keeping only the predicted stickers with the highest confidence in each image + sticker vs. not sticker binary classifier
\end{enumerate}
Both of these methods resulted in very similar F1 scores; Method 1 resulted in higher precision while Method 2 resulted in higher recall.  One method can be chosen over the other based on the need for precision versus the need for recall in the particular use case.  

The test set results of these complete pipelines and the results after each intermediate step are shown in Table \ref{tbl: test res}.  As on the validation set, both complete pipelines (bolded lines in the table) result in very similar F1 scores, with Method 1 having the higher precision and Method 2 having the higher recall. 

\begin{table}
\begin{center}
\begin{tabular}{|c|c|c|c|c|}
\hline
   &
   \begin{tabular}[c]{@{}c@{}}Prediction \\Pipeline\end{tabular} &  \begin{tabular}[c]{@{}c@{}}Precision\end{tabular}
   &\begin{tabular}[c]{@{}c@{}}Recall\end{tabular}&\begin{tabular}[c]{@{}c@{}}F1 Score\end{tabular}\\
\hline
\hline
\multirow{7.25}{*}{\rotatebox{90}{1664 x 1664 images}} & 
\begin{tabular}[c]{@{}c@{}}YOLOv2 object \\detector only\end{tabular} & 0.6068 & 0.8208 & 0.6978 \\
\cline{2-5}
&\begin{tabular}[c]{@{}c@{}}\textbf{YOLOv2 object} \\ \textbf{detector} \\ \textbf{+} \\ \textbf{Keeping only} \\ \textbf{highest confidence} \\ \textbf{predictions} \end{tabular} & \textbf{0.6157} & \textbf{0.8150} & \textbf{0.7015} \\
\hline 
\multirow{16}{*}{\rotatebox{90}{1248 x 1248 images}} & 
\begin{tabular}[c]{@{}c@{}}YOLOv2 object \\detector only\end{tabular} & 0.3282 & 0.9769 & 0.4913 \\
\cline{2-5}
&\begin{tabular}[c]{@{}c@{}} YOLOv2 object \\detector \\+ \\Keeping only \\ highest confidence \\predictions\end{tabular} & 0.3394 & 0.9769 & 0.5037 \\
\cline{2-5}
&\begin{tabular}[c]{@{}c@{}}\textbf{YOLOv2 object} \\\textbf{detector} \\ \textbf{+} \\ \textbf{Keeping only} \\ \textbf{highest confidence} \\ \textbf{predictions}\\ \textbf{+} \\ \textbf{Sticker vs. not} \\ \textbf{sticker classifier}\end{tabular} &  \textbf{0.5668} & \textbf{0.9075} & \textbf{0.6978}\\
\hline
\end{tabular}
\end{center}
\caption{Test set results for best detection pipelines.  Final results of pipelines are bolded; intermediate steps are shown in other rows.   }
\label{tbl: test res}
\end{table}

%------------------------------------------------------------------------
\section{Future work}
Several potential further improvements to the parking sticker detection will be discussed in this section.  

Changing the detection threshold (the confidence value needed to call something a detected sticker) to be higher may improve results.  Because keeping only the detected sticker with the highest confidence in each image improved precision without affecting recall, it is likely than many false positives have lower confidence values than many true positives.  Tuning this hyperparameter could therefore improve results for the YOLOv2 models discussed above.  

Using an ensemble of the 1248 x 1248 YOLOv2 model and the 1664 x 1664 YOLOv2 model could also improve results.  The 1248 x 1248 model has higher recall, while the 1664 x 1664 model has higher precision.  Ensembling the two, therefore, could improve results by making use of the strengths of both models.  

Replacing the YOLOv2 object detectors with Faster R-CNN models \cite{FasterRCNN} could also improve results because Faster R-CNN tends to perform better than YOLOv2 for small objects.  

Several changes to the problem itself could also improve results.  Collecting and annotating more data would give the models more examples from which to learn and would likely decrease overfitting and improve performance.  Making the parking stickers brightly colored instead of plain white with black numbers would simplify the problem by making them easier to detect, both for the humans annotating the data and for the models.  Changing the camera placement would also likely improve results substantially.  Much of the challenge of this problem comes from the non-ideal camera placement that makes the stickers cover a very small percentage of each image.  Placing a camera above the approaching cars could make the stickers larger and more consistently centered in the images.  This particular camera placement would have the added benefit of being similar to cameras that could be placed above cars at stoplights or on highway signs.  These cameras could enable detection of other objects in and on cars, like cell phones in the hands of drivers; the lessons learned from doing parking sticker detection using cameras with similar positioning could be beneficial in addressing this problem.  

%------------------------------------------------------------------------
\section{Conclusion}
ParkingSticker is a very realistic dataset that closely mimics the type of data that could be available in many industry problems.  The small size of the objects of interest, small size of the dataset, non-ideal camera positioning, and non-ideal lighting are characteristics that are present in many real-world problems but are not present in many existing popular datasets.  The preliminary experimental results presented above indicate that detection of the parking stickers in ParkingSticker is a challenging task.  This dataset can be used as a useful benchmark for comparison of various object detection techniques and can be used to aid in the selection of appropriate techniques for similar industry problems.

{\small
\bibliographystyle{ieee}
\bibliography{psbib}
}

\end{document}